\documentclass{article}
\usepackage{spconf,amsmath,graphicx,hyperref}
\usepackage{algorithmic}
\usepackage{algorithm}
\usepackage{array}
\usepackage[caption=false,font=normalsize,labelfont=sf,textfont=sf]{subfig}
\usepackage{textcomp}
\usepackage{stfloats}
\usepackage{url}
\usepackage{verbatim}
\usepackage{booktabs}
\usepackage{cite}
\usepackage{graphicx, wrapfig}
\usepackage{algorithm, algorithmic}
\usepackage{amssymb}
\usepackage{pifont}
\usepackage{multirow}
\usepackage{bbding}

\usepackage{booktabs}
\usepackage{orcidlink}
\usepackage{multirow}

\title{\vspace{-1cm}DGE-YOLO: Dual-Branch Gathering and Effiicient Attention for Accurate UAV Object Detection}
\twoauthors
 {A. Kunwei Lv, B. Zhiren Xiao, C. Hang Ren, D. Ping Lan\sthanks{Thanks to National Key R\&D Program (grant number 2023YFC3321705).}}
	{Xizang University. A-D\\
	 Lhasa, China A-D}
 {E. Xiali Li}
	{Minzu University of China E\\
	Beijing,China E}
\begin{document}
%
\maketitle
\begin{abstract}
The rapid proliferation of unmanned aerial vehicles (UAVs) has highlighted the importance of robust and efficient object detection in diverse aerial scenarios. Detecting small objects under complex conditions, however, remains a significant challenge.To address this, we present DGE-YOLO, an enhanced YOLO-based detection framework designed to effectively fuse multi-modal information. We introduce a dual-branch architecture for modality-specific feature extraction, enabling the model to process both infrared and visible images. To further enrich semantic representation, we propose an Efficient Multi-scale Attention (EMA) mechanism that enhances feature learning across spatial scales. Additionally, we replace the conventional neck with a Gather-and-Distribute(GD) module to mitigate information loss during feature aggregation. Extensive experiments on the Drone Vehicle dataset demonstrate that DGE-YOLO achieves superior performance over state-of-the-art methods, validating its effectiveness in multi-modal UAV object detection tasks.
\end{abstract}
\section{Introduction}
\label{sec:intro}

Unmanned aerial vehicle object detection~\cite{tang2024fourier,liu2024feature} has become a critical component in aerial perception systems. With increasing demand for real-time and high-accuracy detection in complex aerial scenarios, challenges such as small object localization, modality imbalance, and computational constraints become prominent. Recent advances in image generation and multi-modal learning have demonstrated the benefits of modality-specific modeling and information consistency~\cite{weng2024enhancing,li2024lr,qiao2022novel}. Inspired by these insights, we explore how the complementary advantages of infrared and visible images can be harnessed to improve UAV-based object detection. Instead of treating the fusion of modalities as a preprocessing step, we propose a novel lightweight multi-modal detection framework, termed {DGE-YOLO}, that performs joint feature extraction, attention-guided fusion, and global context aggregation in an end-to-end manner.

\begin{figure}[!t]
    \centering
    \includegraphics[width=0.9\linewidth]{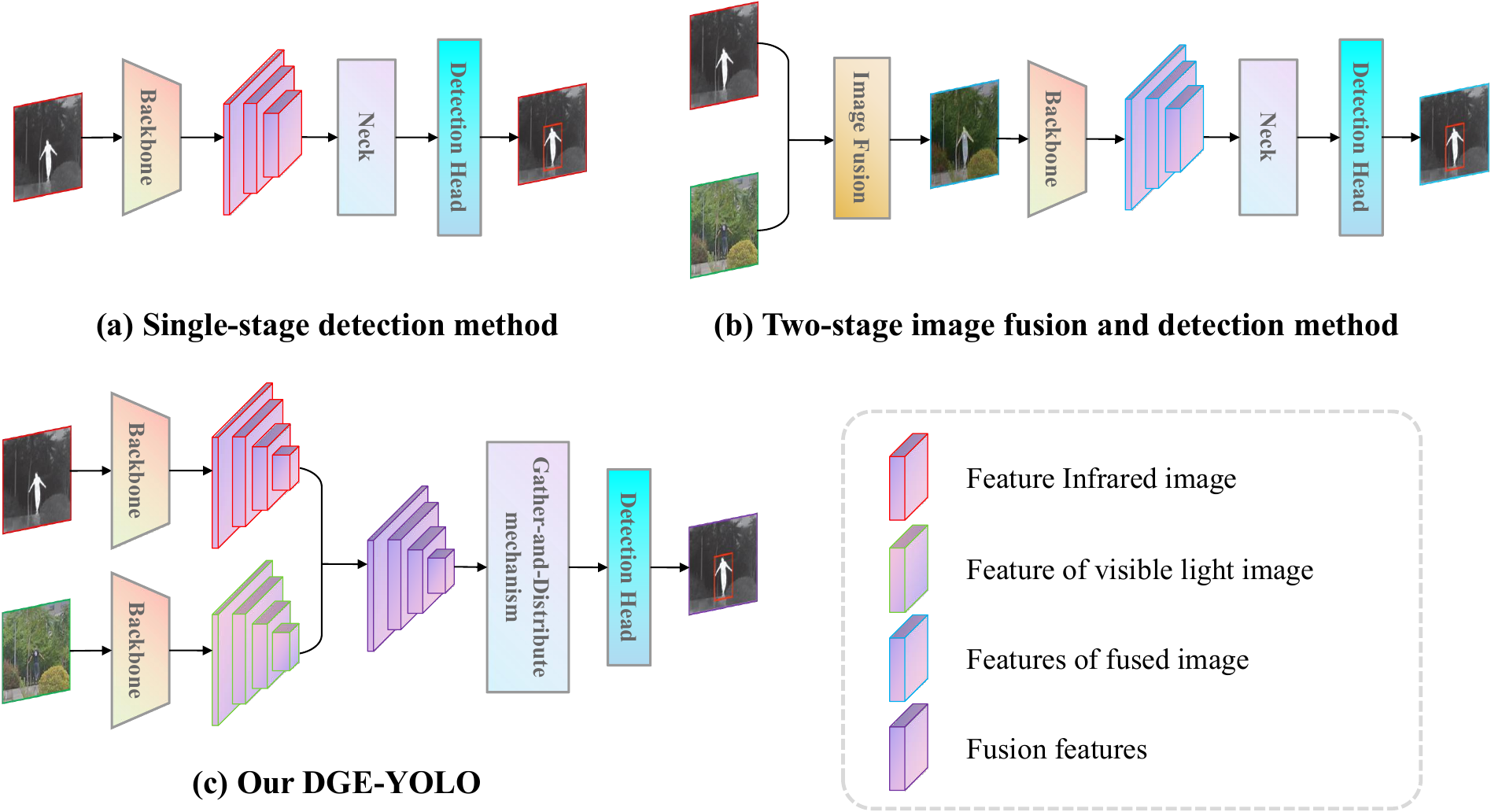}
    \caption{Schematic diagrams of three types of detection frames. }
    \vspace{-0.5cm}
\label{fig:1}
\end{figure}
\begin{figure*}[!t]
    \centering
    \vspace{-1cm}\includegraphics[width=0.9\linewidth]{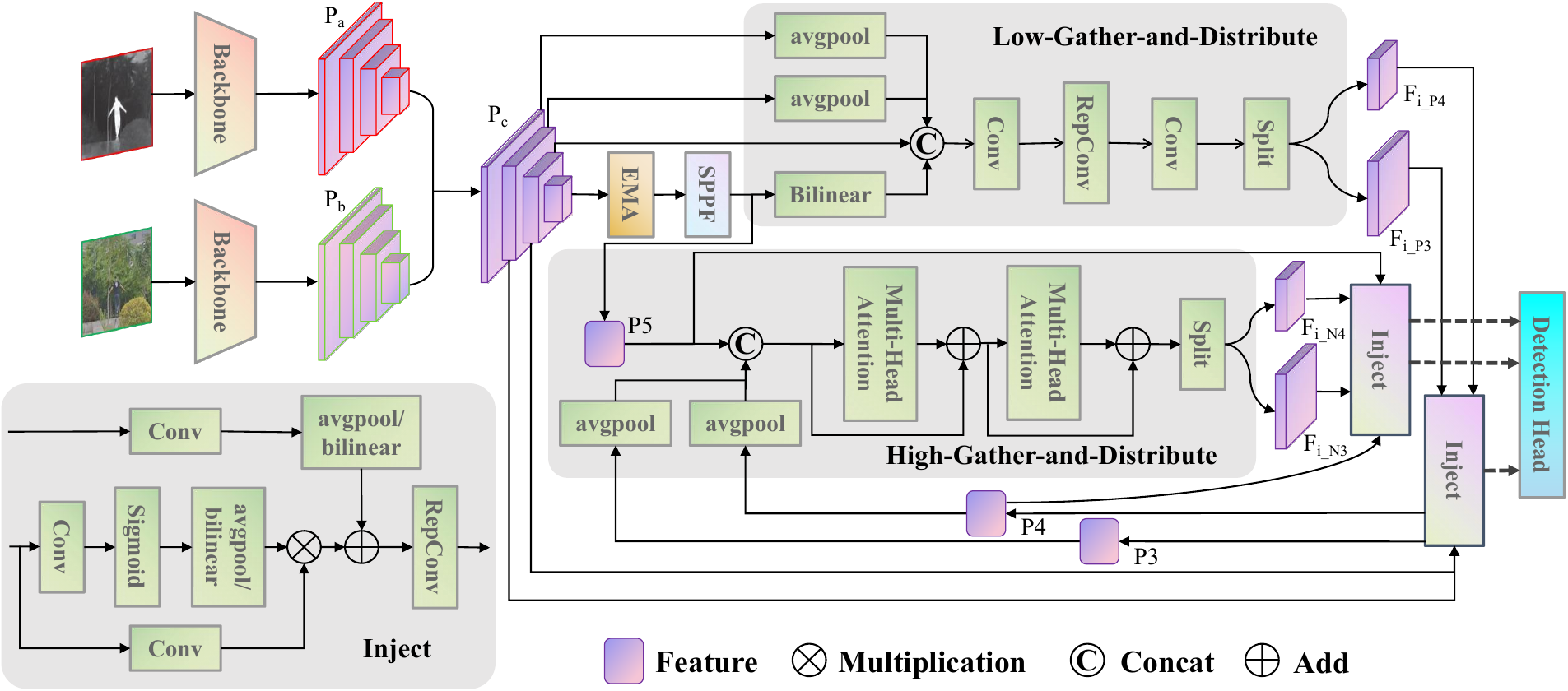}
\caption{The architecture of the proposed DGE-YOLO. Where $P_a$ and $P_b$ respectively represent Feature extracted from infrared and visible images by backbone, while $P_c$ represents Feature after fusion. $P_c$ is divided into $B2$, $B3$, $B4$ and $B5
$ for subsequent processing. Avgpool, Billinear and concat in Low-Gather-and-Distribute form Low-FAM, and the rest is Low-IFM. Similarly, avgpool and concat in High-Gather-and-Distribute form High-FAM, and the rest is High-IFM.}
\label{fig:2}
\end{figure*}

To address these issues, our proposed DGE-YOLO introduces three core improvements. First, we design a dual-branch backbone that processes infrared and visible inputs separately and fuses them across multiple semantic scales. This design draws upon the hierarchical conditioning principles explored in prior conditional generation frameworks~\cite{ shen2025imaggarment}. Second, we incorporate an EMA module~\cite{2023Efficient} to dynamically model long-range dependencies and enhance cross-scale feature representation. Third, we integrate a GD mechanism~\cite{NEURIPS2023_a0673542} as the neck module, allowing global information to be aggregated and redistributed effectively across the feature hierarchy.
Fig~\ref{fig:1}. illustrates the differences between traditional image-level fusion, existing two-stage models, and our proposed end-to-end framework. Extensive experiments on the DroneVehicle~\cite{9759286} dataset show that DGE-YOLO achieves superior detection performance under various aerial scenarios.
Our contributions can be summarized as follows:
We propose a dual-branch lightweight backbone for multi-modal UAV object detection, which processes infrared and visible streams independently before multi-level fusion, balancing accuracy and efficiency.
We introduce a EMA to enhance both local and global feature dependencies across spatial resolutions.
We adopt a GD mechanism as the neck to enable efficient cross-layer feature aggregation and semantic propagation.

\section{METHOD}
\label{sec:pagestyle}
\subsection{Overview}
To address the challenges faced by drones in performing target detection tasks in complex scenarios, such as slow detection speed and high missed detection rates, this paper modifies the backbone network of the YOLOv8n model by implementing a dual-channel input and introducing the EMA self-attention mechanism. Additionally, the neck network structure is replaced with the GD mechanism to further enhance the model's performance. As shown in Fig.\ref{fig:2}
\subsection{Dual-Branch Feature Extraction Network} 
Feature fusion based on multimodal images generally divided into three strategies: early fusion, middle fusion, and late fusion. Middle fusion involves feature fusion at specific feature layers, which avoids the shortcomings of early fusion, such as the inability to effectively handle high-level features, while preserving the independent information of each modality. It also addresses the issue of late fusion not being able to effectively capture the relationships between different layers. Compared to the original model, which outputs from three feature extraction layers, we concatenate the features from all four feature extraction layers, ensuring that the network can fully utilize features from different layers, thereby enhancing the model's ability to handle multi-scale tasks.
\subsection{Effiicient Multi-Scale Attention Module} 
\begin{figure}[!t]
    \centering
    \includegraphics[width=7cm,height=4cm]{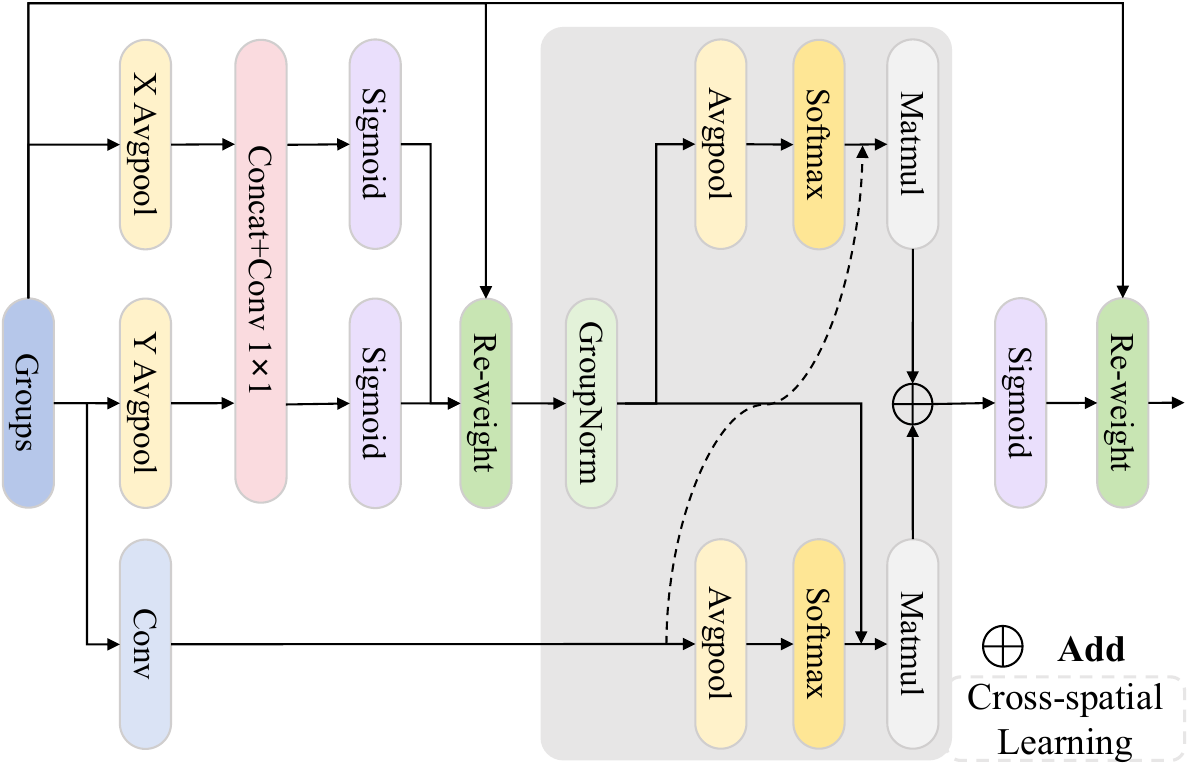}
       \vspace{-0.5cm}
\caption{EMA Module Network Architecture Diagram.}
   \vspace{-0.5cm}
\label{fig:3}
\end{figure}
Effiicient multi-Scale attention module avoids deeper sequential processing and large depth by utilizing parallel substructures. While compressing noise in the process EMA uses three parallel paths to extract attention weight descriptors for the grouped feature maps.  The output from the $1\times1$ branch is encoded as global spatial information through two-dimensional global average pooling, while the output from the $3\times3$ branch is directly transformed into the corresponding dimensional shape. These outputs are then aggregated via matrix dot product operations to generate a spatial attention map. Finally, the attention weights generated by the Sigmoid function are applied to aggregate the output feature maps within each group, capturing pixel-level pairwise relationships and emphasizing the global context of all pixels. The final output maintains the same size as the input, allowing it to be effectively stacked within the network structure. Through the parallel sub-networks, EMA can adjust the weights of different channels by encoding cross-channel information and precisely retain spatial structural information within the channels. EMA also enriches feature aggregation by providing a cross-space information aggregation method in different spatial dimensions.The overall architecture is shown in Fig.\ref{3}.

\subsection{Gather-and-Distribute Mechanism}
\begin{table*}
\renewcommand\arraystretch{1}
\centering
\vspace{-0.5cm}
\belowrulesep=0pt
\aboverulesep=0pt
\caption{Comparison with State-of-the-art Methods.THE {\color{red}RED}, {\color{green}GREEN} AND {\color{blue}BLUE} ROWS REPRESENT THE 1ST, 2ND AND 3RD PLACES.RESPECTIVELY.} 

\resizebox{15cm}{!}
{
\begin{tabular}{c|c|cc|ccccc|cc}
   
   \toprule
        Detectors&Modality&Param(M)& Flops(G)&Car&Truck&Freight car&Bus& Van&mAP50(\%)&mAP50:95(\%)\\
   
   \midrule
        RetnaNet& \multirow{10}{*}{RGB}  & 36.5 & 146.2 & 78.5 & 34.4 & 24.1 & 69.8 & 28.8 & 47.1 & -  \\ 
        Faster R-CNN&  & 41.5 & 192.6 & 79.9 & 49.0 & 37.2 & 77.0 & 37.0 & 55.9 & -   \\ 
        Oriented R-CNN\cite{9710901}&  & 43.0 & 241.4 & 80.1 & 53.8 & 41.6 & 85.4 & 43.3 & 60.8 & -   \\ 
        $s^2$A-Net\cite{9377550}& & 48.7 & 286.2 & 80.0 & 54.2 & 42.2 & 84.9 & 43.8 & 61.0 & -  \\
        RoITransformer\cite{8953881}&  & 53.4 & 336.4 & 61.6 & 55.1 & 42.3 & 85.5 & 44.8 & 61.6 & - \\ 
        YOLOv8s\cite{yolov8_ultralytics}&  & 11.1 & 28.4 & {\color{blue}96.1} & 74.6 & 62.1 & 95.6 & 63.9 & 78.4 & 56.1 \\ 
        YOLOv9s\cite{wang2024yolov9} &  & {\color{green}7.3} & 26.7 & {\color{blue}96.1} & 76.0 & 64.4 & 96.2 & 64.8 & 79.5 & 57.6 \\ 
        YOLOv10s\cite{THU-MIGyolov10} &  & {\color{blue}8.0} & 24.5 & 96.0 & 74.8 & 61.6 & 95.9 & 64.8 & 78.6 & 56.5 \\ 
        YOLOv11s\cite{yolo11_ultralytics} &  & 9.4 & {\color{blue}21.3} & 96.0 & 75.7 & 60.4 & 96.3 & 64.4 & 78.5 & 56.5 \\  
        YOLOv12s\cite{yolo12} &  & 9.2 & {\color{green}21.2} & 96.0 & 76.0 & 63.9 & 95.9 & 63.4 & 79.0 & 56.7 \\  
        \midrule
        RetnaNet& \multirow{10}{*}{IR} & 36.5 & 146.2 & 88.8 & 35.4 & 39.5 & 76.5 & 32.1 & 54.5 & -  \\ 
        Faster R-CNN&  & 41.5 & 192.6 & 89.4 & 53.5 & 48.3 & 87.0 & 42.6 & 64.2 & -   \\ 
        Oriented R-CNN&  & 43.0 & 241.4 & 89.8 & 57.4 & 53.1 & 89.3 & 45.4 & 67.0 & -   \\ 
        $s^2$A-Net&  & 48.7 & 286.2 & 89.9 & 54.5 & 55.8 & 88.9 & 48.4 & 67.5 & -  \\
        RoITransformer&  & 53.4 & 336.4 & 90.1 & 60.4 & 58.9 & 89.7 & 52.2 & 70.3 & - \\ 
        YOLOv8s&  & 11.1 & 28.4 & {\color{blue}{\color{green}98.4}} & 76.8 & 73.4 & 97.1 & 66.1 & 82.4 & 61.6 \\ 
        YOLOv9s &  & {\color{green}7.3} & 26.7 & {\color{red}98.5} & {\color{blue}79.2} & {\color{blue}75.7} & {\color{red}97.4} & {\color{green}68.4} & 83.8 & {\color{green}63.3} \\ 
        YOLOv10s &  & {\color{green}8.0} & 24.5 & {\color{red}98.5} & 78.6 & 73.5 & 94.9 & 65.9 & {\color{blue}84.1} & {\color{blue}62.9} \\ 
        YOLOv11s &  & 9.4 & {\color{blue}21.3} & {\color{blue}{\color{green}98.4}} & {\color{green}80.0} & {\color{red}77.4} & {\color{blue}97.0} & {\color{blue}67.9} & {\color{green}84.2} & {\color{blue}62.9} \\  
        YOLOv12s &  & 9.2 & {\color{green}21.2} & {\color{blue}{\color{green}98.4}} & 78.2 & 73.8 & {\color{green}97.2} & 67.3 & 83.0 & 62.3 \\ 
        \midrule
        UA-CMDet &\multirow{10}{*}{RGB+IR}& - & - & 87.5 & 60.7 & 46.8 & 87.1 & 38.0 & 64.0 & -  \\ 
        Halfway Fusion\cite{BMVC2016_73} &  & - & - & 90.1 & 62.3 & 58.5 & 89.1 & 49.8 & 70.0 & -   \\ 
        CIAN\cite{ZHANG201920} &  & - & - & 90.1 & 63.8 & 60.7 & 89.1 & 50.3 & 70.8 & -    \\ 
        AR-CNN\cite{9523596} &  & - & - & 90.1 & 64.8 & 62.1 & 89.4 & 51.5 & 71.6 & -   \\ 
        MBNet\cite{10.1007/978-3-030-58523-5_46}&  & - & - & 90.1 & 64.4 & 62.4 & 88.8 & 53.6 & 71.9 & -  \\
        TSFADet\cite{10.1007/978-3-031-20077-9_30}&  & - & - & 89.9 & 67.9 & 63.7 & 89.8 & 54.0 & 73.1 & - \\ 
        $c^2$Former\cite{10472947}&  & -& - & 90.2 & 68.3 & 64.4 & 89.8 & 58.5 & 74.2 & - \\ 
        SLBAF-Net\cite{10.1007/s11042-023-15333-w} &  & - & - & 90.2 & 72.0 & 68.6 & 89.9 & 59.9 & 76.1 & - \\ 
        OAFA &  & - & - & 90.3 & 76.8 & 73.3 & 90.3 & 66.0 & 79.4 & - \\ 
        Ours &  & {\color{red}7.1} & {\color{red}15.2} & {\color{red}98.5} & {\color{red}80.8} & {\color{green}77.1} & {\color{green}97.2} & {\color{red}69.9} & {\color{red}84.7} & {\color{red}66.1} \\ 
   \bottomrule
\end{tabular}
}
\label{table:1}
\end{table*}

\noindent
\textbf{Stage Feature Alignment Module.} In the low-stage feature alignment module (Low-FAM), input features are downsampled using average pooling, and the features are adjusted to the smallest feature size within the group ($R_{B4}$ = 1/4$R$, $R$ represent size) to obtain aligned features. In the high-stage feature alignment module (High-FAM), average pooling is used to adjust the dimensions of the input features to a uniform size of $R_{B5}$ = 1/8$R$. The feature alignment module ensures efficient aggregation of information while minimizing the computational complexity of subsequent.

\noindent
\textbf{Stage Feature Fusion Module.} The low-stage information fusion module (Low-IFM) includes multi-layer reparameterized convolution blocks and segmentation operations. The results obtained from Low-FAM undergo a RepBlock operation to generate $F_{low\_ fuse}$, which is then split along the channel dimension using the segmentation operation into $F_{i\_P3}$ and $F_{i\_ P4}$. These are then fused with features from different layers. Similarly, the High-Stage Information Fusion Module(High-IFM) contains Transformer blocks and segmentation operations. The results from High-IFM are passed through the Transformer block to generate $F_{high\_fuse}$, which is then reduced in the number of channels via convolution to sum($C_{P4}$, $C_{P5}$). Finally, the channels are split into $F_{i\_N4}$ and $F_{i\_N5}$, which are fused with the current layer's features for integration.

\noindent
\textbf{Information Injection Module.}
The Information injection module first takes the local features $F_l$ from the current layer and the injected information $F_i$ generated by the IFM  as inputs. It then applies two different convolutional layers to $F_i$ to compute $F_{g\_embed}$ and $F_{act}$. These two processed features are then fused using an attention mechanism to produce the final output feature $F_{out}$. To ensure that the dimensions of local features $F_l$ and global features$F_g$ (the combined features from the current layer and the injected information) align with the size of $F_i$, average pooling or bilinear interpolation is used to scale $F_{embed}$ and Fact appropriately. Finally, RepBlock is applied to further extract and refine the fused information. 
\begin{equation}\label{1}
F_{g\_act}=resize(Sigmoid(Conv_{act}(F_i)).
\end{equation}
\begin{equation}\label{2}
F_{g\_embed}=resize(Conv_{g\_embed}(F_i)).
\end{equation}
\begin{equation}\label{3}
F_{att_{fuse}}=Conv_{l\_embed}(F_l)*F_{i\_act}+F_{g\_embed}.
\end{equation}


\section{Experiment and Analysis} \label{sec:exp}  
\begin{figure*}[!t]
    \centering
\includegraphics[width=0.8\linewidth]{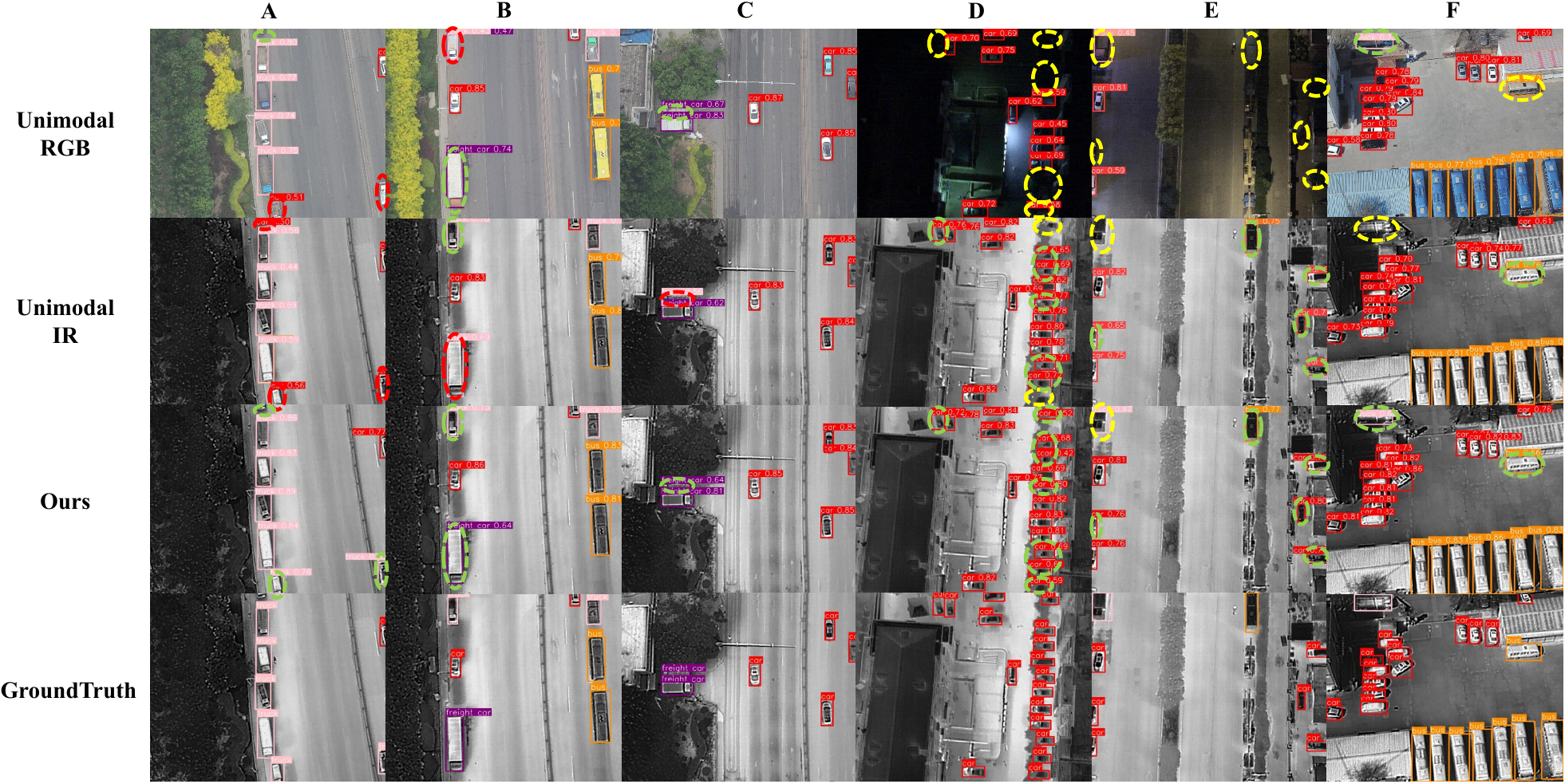}
    \vspace{-0.3cm}
\caption{
Visualization results. Compared our method with results of baseline in two unimodal. The red detection box represents the car class, pink represents the truck class, purple represents the freight car class, and orange represents the bus class.  The red dashed boxes represent misdetected targets, the yellow dashed boxes represent missed targets, and the green dashed boxes represent correctly detected targets.}
    \vspace{-0.2cm}
\label{fig:4}
\end{figure*}

\subsection{Datasets}
\noindent
\textbf{Drone Vehicle.}
This paper evaluates our method on the Drone Vehicle dataset, The dataset includes five object categories: car, truck, bus, freight car, and van. Following the training approach in the original paper, we used 17,990 image pairs for training and 1,469 image pairs for validation. 

\subsection{Comparison with State-of-the-art Methods}

The experimental results are shown in Table\ref{table:1}. Among the single-modal detection models, the detection performance under the infrared modality is significantly higher than that under the visible light modality. We selected the YOLOv8 and subsequent versions to validate the effectiveness of our model. Compared to previous detection models, our model exhibits a significant improvement in accuracy. Since the parameters and flops of the single-stage detection model are much smaller than those of the multimodal fusion model, no comparison is made in the experiment. Our model has shown substantial improvements in categories with previously lower accuracy, achieving the highest mAP50, and a significant boost in the mAP50:95 metric.This paper proposed DGE-YOLO method outperforms many state-of-the-art single-modal and multimodal object detection methods.

\begin{table}
\renewcommand\arraystretch{1}
\centering
\vspace{-0.8cm}
\belowrulesep=0pt
\aboverulesep=0pt
\caption{Ablation Studies.} 
\resizebox{8cm}{!}
{
\begin{tabular}{c|cccc|cc}
   
   \toprule
        ID&RGB&IR&EMA&GD&mAP50(\%)&mAP50:95(\%)\\
   
   \midrule
        A& \checkmark & \XSolidBrush & \XSolidBrush& \XSolidBrush & 74.1 & 52.5 \\
        B& \XSolidBrush & \checkmark& \XSolidBrush &  \XSolidBrush & 74.6 & 52.5 \\
        C& \checkmark & \checkmark & \XSolidBrush& \XSolidBrush & 84.2& 63.4\\
        D& \checkmark & \checkmark & \checkmark &  \XSolidBrush & 83.0& 65.2\\
        E& \checkmark& \checkmark & \XSolidBrush & \checkmark & 84.3& 63.2\\
        F& \checkmark & \checkmark & \checkmark & \checkmark & \textbf{84.7}& \textbf{66.1}\\

   \bottomrule
\end{tabular}
}
\label{table:2}
\end{table}

\begin{figure}[!t]
    \centering
\setlength{\abovecaptionskip}{-0.5cm}
 
    \includegraphics[width=\linewidth]{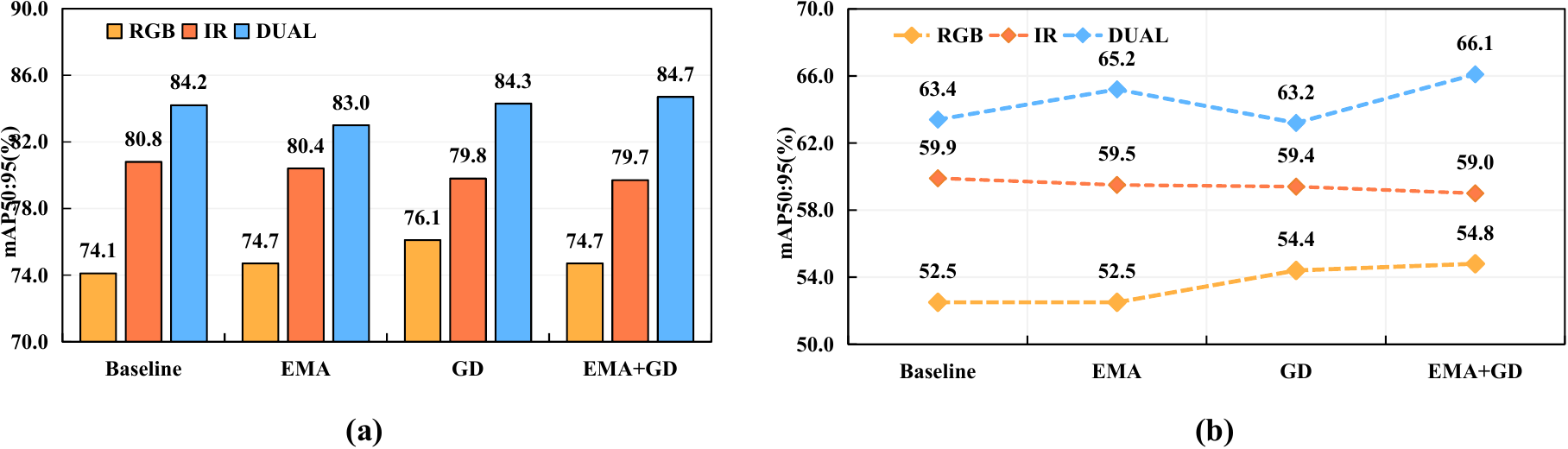}
    \vspace{-0.5cm}
\caption{Introduce comparisons of different modules in different modalities.}
    \vspace{-0.8cm}
\label{fig:5}
\end{figure}

\noindent\textbf{Visualization.}  
In this section, we validate the effectiveness of the model through visual demonstrations. Since image fusion has not been applied, we chose the infrared modality, which contains more feature information, to represent our model for performance display. As shown in Fig.\ref{fig:4}, under dark conditions, many targets are difficult to detect in the visible light modality, resulting in a significant loss of targets. From the detection results, it can be observed that the original model has different advantages when detecting the same target in the visible light and infrared modalities. Overall, the proposed model not only addresses the issue of missed detections effectively but also improves the recognition of false positives.
\noindent\textbf{Ablation Studies.} 
In this section, we conduct ablation experiments to validate the effectiveness of each module in the model. The experimental results are presented in TABLE\ref{table:2}. The detection accuracy of the bimodal approach is higher than both the infrared and visible light single-modality detection accuracies, This indicates that the infrared modality contains more feature information than the visible light modality.while the EMA module performs better in improving mAP50:95, As shown in Fig.\ref{fig:5}(a). Because the module performs better in dealing with complex fusion Feature, it also performs generally when single mode is introduced. The experimental results show that the Gather-and-Distribute module is more effective in improving mAP50,  this module effectively improves the overall performance of the model, As shown in Fig.\ref{fig:5}(b).

\section{Conclusion}\label{sec:con} 
We presented DGE YOLO, a multi modal UAV detector that combines a dual branch backbone for infrared and visible inputs, an Efficient Multi scale Attention module to strengthen cross scale semantics, and a Gather and Distribute neck that reduces aggregation loss. Experiments on the Drone Vehicle dataset show consistent gains over strong baselines and prior state of the art, particularly for small objects in complex scenes. The framework is compatible with existing YOLO variants and offers a practical solution for robust and efficient aerial perception.

{\small
\bibliographystyle{IEEEbib}
\bibliography{strings,refs}
}
\end{document}